\documentclass[10pt,twocolumn,letterpaper]{article}
\usepackage{cvpr}

\definecolor{cvprblue}{rgb}{0.21,0.49,0.74}
\usepackage[pagebackref,breaklinks,colorlinks,allcolors=cvprblue]{hyperref}

\usepackage{graphicx}
\usepackage{amsmath}
\usepackage{float}
\usepackage{xcolor}
\usepackage{lipsum}

\def\paperID{*****}
\def\confName{CVPR}
\def\confYear{2026}

\title{The Impact of VAE Design on Latent Pose Representations for Diffusion-based Sign Language Production}

\author{Guilhem Fauré, Mostafa Sadeghi, Sam Bigeard, Slim Ouni\\
Université de Lorraine, CNRS, Inria, LORIA\\
F-54000 Nancy, Grand Est, France\\
{\tt\small \{guilhem.faure, mostafa.sadeghi, sam.bigeard\}@inria.fr, slim.ouni@loria.fr}
}

\begin{document}
\maketitle
\begin{abstract}

Latent diffusion approaches to sign language production (SLP) rely on an initial stage that learns an encoding of sign pose sequences, enabling generative modeling in the resulting latent space. The autoencoder used in this stage is typically evaluated in terms of reconstruction quality using geometric metrics common in SLP. While informative, these metrics do not fully capture latent space properties that may influence the training and performance of the downstream generative model. In this work, we investigate how architectural and training objective design choices in a variational autoencoder (VAE) for sign pose encoding affect latent space structure, and how these differences translate into the performance of a latent diffusion model for text-to-sign generation. Our experiments on \textsc{Phoenix14T} dataset show that variations in generative performance, measured through back-translation BLEU scores, can sometimes be better explained by differences in latent space properties than by VAE reconstruction accuracy alone.

\end{abstract}
\section{Introduction}
\label{sec:introduction}

In the last years, deep learning methods have driven progress in different sign language processing tasks, notably sign language recognition and translation \cite{slr_1, slt_1}. More recently, sign language production (SLP), referring to the generation of sign language sequences from spoken or written language, has received increasing attention \cite{review_slp}.

Early deep neural SLP work primarily relied on autoregressive architectures, notably with the Transformer encoder-decoder baseline from \citet{saunders_prog_trans}. In parallel, diffusion models have emerged as a promising alternative \cite{xie_gloss_discrete_diffusion, baltatzis_neural_sign_actors, he_text_driven_diff}, offering improved robustness to classical issues of autoregressive models such as error accumulation or regression-to-the-mean. Among these approaches, latent diffusion models have proven particularly effective for conditional image generation~\cite{Rombach_2022_CVPR} and have since been successfully extended to other modalities, e.g., audio \cite{audio_ldm} or video \cite{magic_video}. For SLP in particular, a recent study from \citet{feng_text2sign_latentdiff} explored (continuous) latent diffusion and reported state-of-the-art performance over previous approaches on German Sign Language (DGS) and American Sign Language (ASL) benchmark datasets in terms of standard SLP geometric and back-translation metrics.

A key component of latent diffusion models is the latent representation on which the diffusion process operates. In practice, this representation is typically obtained using a a variational autoencoder (VAE) \cite{KingW14} encoding input sequences into a compact latent space and reconstructing them from it. The design of such autoencoders is commonly guided by reconstruction accuracy. However, prior works in image generation \cite{yao_reconstruction_vs_gen, yao_pretraining_generation} suggest that optimizing solely for reconstruction error does not necessarily yield the most suitable latent representations for generative modeling.

Motivated by this observation, we investigate the impact of VAE design on the latent sign pose representations for diffusion-based SLP. More precisely, we study how different architectural and training choices for the sign pose VAE affect the properties of the learned latent space, and how the latter relates to the performance of a latent diffusion SLP model. To the best of our knowledge, such an analysis has not yet been explored in the context of diffusion-based SLP.

\noindent Our main contributions can be summarized as follows:
\begin{itemize}
    \item We design and evaluate four VAE variants for encoding sign language pose sequences. Implementations of these VAEs are made available online\footnote{\url{https://github.com/GFaure9/SignPoseVAE}}.
    
    \item We analyze how these design choices impact latent space properties and reconstruction performance.
    
    \item We analyze how the VAE reconstruction and latent space metrics relate to latent diffusion SLP performance, highlighting emerging trends and providing intuitive interpretations to build insights for the design of VAEs in diffusion-based SLP.
\end{itemize}
\section{Related Work}
\label{sec:related_work}

\subsection{Sign Language Production}

Various deep neural SLP approaches are based on autoregressive models, in which sign poses are generated sequentially \cite{saunders_prog_trans, yin_vq_autoregressive_gpt, xie_latent_motion_transformer, walsh_data_driven}. While effective and appropriate for variable length sequential data, this paradigm is prone to challenges such as error propagation during generation and the tendency to produce averaged motions with limited diversity.

To address these limitations, recent works in SLP have explored non-autoregressive formulations, including diffusion approaches \cite{he_text_driven_diff, tang_signidd, baltatzis_neural_sign_actors, tang_gloss_diffusion, xie_gloss_discrete_diffusion, feng_text2sign_latentdiff}, and non-autoregressive architectures based on transformers or spatial-temporal graph convolutional networks \cite{tasyu_disentangle, huang_nat_ea}. They aim to generate pose sequences in a single pass or via iterative refinement, rather than sequentially.

Moreover, independently of whether the generation model is autoregressive or not, many recent SLP systems adopt a two-stage pipeline: first encoding sign pose sequences into compact latent representations through an autoencoder, and then training a generation model in this latent space \cite{xie_latent_motion_transformer, feng_text2sign_latentdiff, walsh_data_driven, yin_vq_autoregressive_gpt}. In several works, this encoding is achieved via vector quantization techniques such as VQ-VAE or VQ-GAN, leading to a learned discrete codebook of sign tokens. While convenient for sequential token prediction, discretization can introduce limitations such as unnatural transitions or the loss of fine-grained motion details.

To mitigate these issues, \citet{tasyu_disentangle} proposed learning continuous sign pose representations through an autoencoder and generating the latent vectors using a non-autoregressive transformer conditioned on textual embeddings. However, deterministic decoders trained with regression losses may still produce over-smoothed motions when multiple plausible trajectories exist \cite{hpgan}. In this context, the latent diffusion approach employed by \citet{feng_text2sign_latentdiff} appears as a promising alternative. In fact, since diffusion explicitly tries to approximate the probability distribution of sign pose latent representations, it might be better suited to the stylistic and temporal variations inherent to sign language motion. The authors report improved performance compared to previous approaches, which underscores the potential of latent diffusion in continuous space for SLP.

\subsection{Latent Diffusion and Latent Representations}

Latent diffusion models, introduced by \citet{Rombach_2022_CVPR} for text-to-image generation, consists of two stages. First, an autoencoder ($\mathcal{E}$, $\mathcal{D}$), typically a VAE, is trained to encode the data $x$ into latent variables $z=\mathcal{E}(x)$. Secondly, a denoising diffusion probabilistic model (DDPM) is trained in this latent space, gradually corrupting $z$ in a forward noising process $q(z^{(1:N_\text{diff})}\mid z^{(0)})
= \prod_{n=1}^{N_\text{diff}} q(z^{(n)} \mid z^{(n-1)})$, and optimizing a neural network $p_{\theta}$ to approximate the reverse denoising process under the condition $c$:
\begin{equation}
p_\theta(z^{(0:N_\text{diff})})
= p(z^{(N_\text{diff})})
\prod_{n=1}^{N_\text{diff}} p_\theta(z^{(n-1)} \mid z^{(n)}, c),
\label{eq:denoising}
\end{equation}
\noindent where $z^{(N_\text{diff})}$ is sampled from a standard normal distribution. Once an estimation of $z^{(0)}$ is recovered at the end of the reverse process, it is then decoded into the data space as $\hat{x} = \mathcal{D}(\hat{z}^{(0)})$.

SLP models that learn continuous latent pose representations in a two-stage pipeline typically employ autoencoders that encode body regions separately (e.g., torso and arms, hands, and face) \cite{tasyu_disentangle, feng_text2sign_latentdiff}. Beyond this common architectural choice, however, there is currently no standardized design for sign pose autoencoders intended for SLP. For instance, \citet{feng_text2sign_latentdiff} employ a variational encoder with mean squared reconstruction losses, whereas \citet{tasyu_disentangle} use deterministic linear encoders with region-weighted $\ell_1$ reconstruction objectives, and a regularization term promoting sparser latent representations.

However, recent work in image generation has shown the strong impact of latent representations on downstream generative performance. In particular, \citet{yao_reconstruction_vs_gen} report that improving image tokenizer reconstruction accuracy by increasing feature dimensionality can significantly degrade generation quality. Similarly, \citet{yao_pretraining_generation} demonstrate that incorporating text-image contrastive and self-supervised objectives during tokenizer training leads to substantial improvements in generative performance.

Therefore, sign pose latent representations might play an important role in two-stage SLP pipelines, which remains insufficiently understood. Notably, the relationship between VAE design choices, characteristics of the resulting latent space, and the performance of downstream diffusion-based sign language generation has not yet been systematically investigated.
\section{Methodology}
\label{sec:method}

\subsection{SLP Latent Diffusion Framework}

For sign language generation, we adopt a conditional latent diffusion model following the framework of \citet{Rombach_2022_CVPR}. More precisely, the diffusion process is performed on learned latent representations of sign pose sequences rather than on the original 3D skeletal pose coordinates, which live in a higher-dimensional space. The latent space originates from a variational encoder (see Figure \ref{fig:sign_pose_vae}) mapping each sequence of $F$ sign poses $x_{1:F}\in\mathbb{R}^{F\times N\times3}$ to a sequence of latent representations $z_{1:F}\in\mathbb{R}^{F \times d_{\text{lat}}}$.

\begin{figure}[hbt!]
\centering
\includegraphics[
    width=\linewidth,
    trim=1cm 0.5cm 1cm 0.5cm
    clip
]{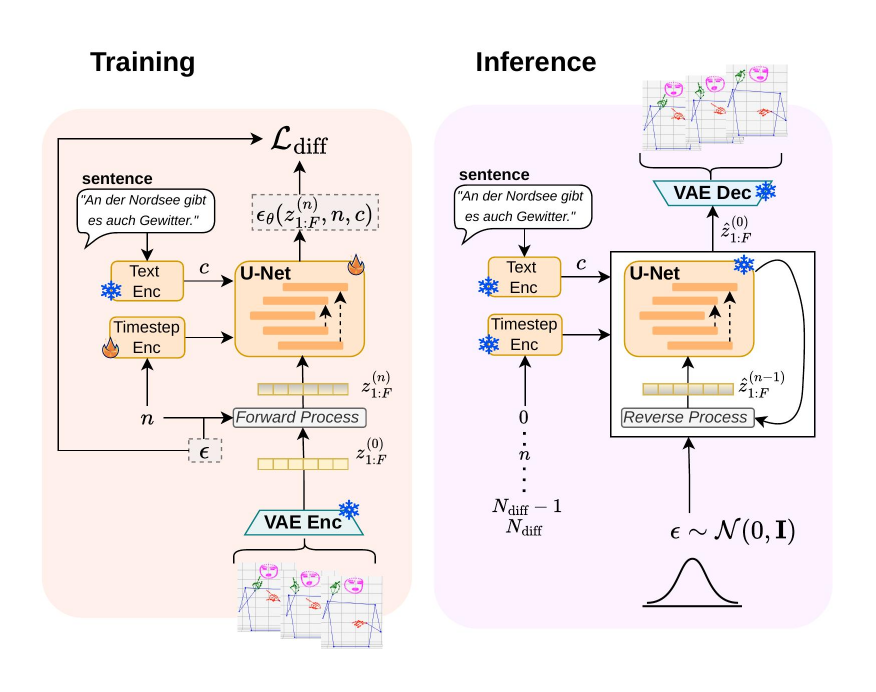}
\caption{Training and inference pipelines of the SLP latent diffusion model. Snowflakes = frozen modules; flames = trainable modules.}
\label{fig:latent_diffusion}
\end{figure}

During training, Gaussian noise is progressively added to the latent representation $z_{1:F}^{(0)}$ according to the following forward diffusion process:
\begin{equation*}
z_{1:F}^{(n)} = \sqrt{\bar{\alpha}_n} z_{1:F}^{(0)} + \sqrt{1 - \bar{\alpha}_n} \epsilon,
\end{equation*}
where $n$ is the diffusion process timestep, $\epsilon \sim \mathcal{N}(0, \textbf{I})$ and $\bar{\alpha}_n = \Pi_{s=0}^n (1 - \gamma_s)$, with $\gamma_s$ a predefined noise schedule.
A multi-resolution U-Net model, denoted $\epsilon_\theta$, is trained to predict the injected noise from the noisy latent $z_{1:F}^{(n)}$ at each timestep (cf. Figure \ref{fig:latent_diffusion}).

To allow for text-conditioned generation of sign pose sequences, the noise prediction at each timestep is conditioned by an embedding $c\in\mathbb{R}^{L\times d_c}$ of the sentence associated to the input noisy latent sign pose sequence. This text embedding consists in token-level features extracted from a pretrained BERT-like language encoder, and is provided as context to the U-Net through cross-attention blocks at several layers.

The training process then amounts to minimizing the following loss:

\begin{equation*}
\mathcal{L}_{\text{diff}} =
\mathbb{E}_{z_{1:F}^{(0)},\epsilon,n}
\left[
\| \epsilon - \epsilon_\theta \left( z_{1:F}^{(n)}, n, c \right) \|_2^2
\right],
\end{equation*}

After denoising (Equation \ref{eq:denoising}), the predicted latent sequence $\hat{z}_{1:F}^{(0)}$ is mapped back to the pose space using the VAE decoder to obtain the generated sign pose sequence.

\subsection{Sign Pose VAE Variants}

\begin{figure*}[t]
\centering
\includegraphics[
    width=\textwidth,
]{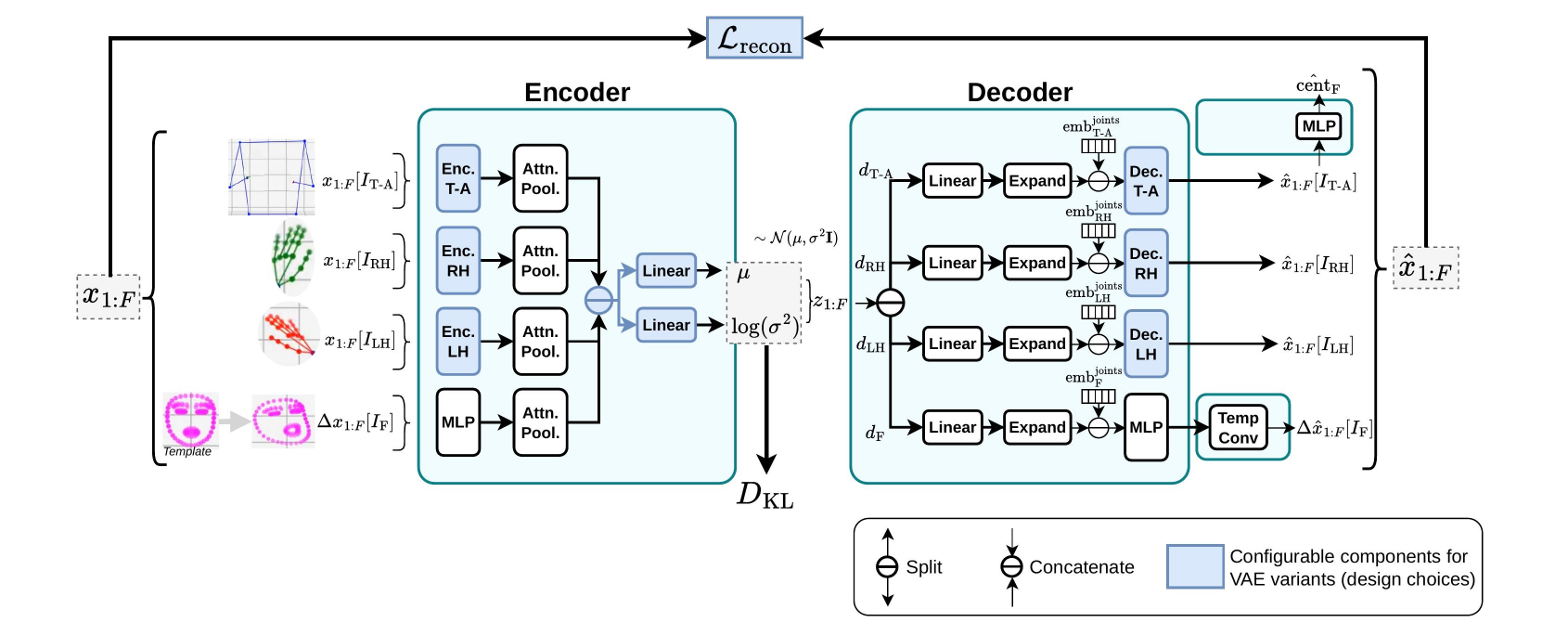}
\caption{Architecture of the Sign Pose VAE. Hands coordinates are centered at their wrists. 
Face coordinates are normalized as 
$\Delta x_{1:F}[I_\text{F}] = (x_{1:F}[I_\text{F}] - \text{cent}_\text{F}) - x^{\text{template}}$, 
where $\text{cent}_\text{F} \in \mathbb{R}^{F \times 1 \times 3}$ denotes the face centroid at each frame and 
$x^{\text{template}} \in \mathbb{R}^{N_\text{F} \times 3}$ is a fixed neutral face template centered at the origin.
$\text{emb}^{\text{joints}}_r \in \mathbb{R}^{N_r \times d_{\text{emb}}}$ denotes learnable joint embeddings for region $r$, used to distinguish joints after broadcasting the shared features across joints ("Expand" operation).
Dark blue = VAE variant-specific components.}
\label{fig:sign_pose_vae}
\end{figure*}

To study the impact of VAE design choices on sign pose latent representations, and consequently on SLP, we consider four VAE variants differing in architecture, reconstruction objectives, and latent distribution.

For all variants, the body is decomposed into four regions: torso and arms (T-A), right hand (RH), left hand (LH), and face (F) (cf. Figure \ref{fig:sign_pose_vae}). Each region is processed by a dedicated encoder and decoder module. By default, the encoded embeddings of the regions are then combined to predict the latent distribution. The latent dimensionality $d_{\text{lat}}=d_{\text{T-A}} + d_{\text{RH}} + d_{\text{LH}} + d_{\text{F}}$, as well as per-region latent dimensions, are kept fixed across all variants to better isolate the effect of variant-specific changes.

\paragraph{General VAE formulation.}
Classically, from an input pose sequence $x_{1:F}$, the encoder first predicts a latent distribution
\begin{equation*}
q_\phi(z_{1:F}|x_{1:F}) = \mathcal{N}(\mu(x_{1:F}), \sigma^2(x_{1:F}) \textbf{I}),
\end{equation*}
\noindent from which a latent sequence $z_{1:F}$ is sampled using the re-parameterization trick, and given to the decoder to reconstruct the pose sequence $\hat{x}_{1:F}$.

To train the VAEs, we employ the training objective from the standard $\beta$-VAE formulation of \citet{Higgins2016betaVAELB}:
\begin{equation*}
\mathcal{L}_{\text{VAE}} =
\mathcal{L}_{\text{recon}} +
\beta \, D_{\mathrm{KL}}\!\left(q_\phi\,||\,\mathcal{N}(0, \textbf{I})\right),
\end{equation*}
\noindent where $\mathcal{L}_{\text{recon}}$ denotes a reconstruction error between the reference and predicted poses, and $D_{\mathrm{KL}}$ is the Kullback–Leibler divergence, acting as a regularization term to encourage the latent distribution to have a Gaussian shape. $\beta$ is progressively increased during training to stabilize reconstruction before stronger KL regularization. \\
In the following, we present the specifics of each variant in more detail.

\paragraph{Variant 1: MLP baseline (\textit{BaseVAE}).}
The first variant serves as the reference baseline. Each region encoder and decoder is implemented as a multilayer perceptron (MLP), while the reconstruction loss consists of an $\ell_1$ error on joint positions averaged over frames for each region:
\begin{equation}
\mathcal{L}_{\text{recon}} =  
\frac{1}{F} \sum_f \left( \sum_{r \in \mathcal{R}} w^{\text{pos}}_r 
\| x_f[r] - \hat{x}_f[r] \|_1
\right)
,
\label{eq:l_recon_v1}
\end{equation}

\noindent where $\mathcal{R}=\{ I_{\text{T-A}}, I_{\text{RH}}, I_{\text{LH}}, I_{\text{F}} \}$ denotes the set of index subsets corresponding to body regions in the skeletal representation, and $w^{\text{pos}}_r$ are per-region weights set to equal values in this variant.

\paragraph{Variant 2: Structured encoder-decoder (\textit{StructVAE}).}
The second variant replaces the MLP modules with architectures that explicitly exploit the skeletal structure and temporal dynamics of pose sequences. Specifically, for the torso/arms and hands regions, we employ graph convolutional (GCN) layers to model the connectivity between joints, followed by residual temporal convolutional (ResTempConv) blocks to capture motion dependencies over time. In principle, we expect this inductive bias to facilitate learning, notably by encouraging latent representations to focus on motion patterns and not having to ''re-discover/re-encode'' joint connectivity. The reconstruction objective is kept as in \textit{BaseVAE}.

\paragraph{Variant 3: Structured model with multi-objective reconstruction (\textit{MultiObjVAE}).}
The third variant builds upon \textit{StructVAE}'s architecture but introduces a richer reconstruction objective. More precisely, we incorporate additional losses that isolate the mouth keypoints position error and penalize the velocity of keypoints for specific regions, namely torso/arms, face, and mouth. Besides, different weighting factors are applied for each region. The reconstruction objective becomes
\begin{equation}
\begin{aligned}
\mathcal{L}_{\text{recon}} &= 
\frac{1}{F} \sum_{f=1}^F \left(
\sum_{r \in \mathcal{R} \cup I_{\text{M}}} w^{\text{pos}}_r
\| x_f[r] - \hat{x}_f[r] \|_1 \right) \\
&+ \frac{1}{F-1} \sum_{f=1}^{F-1} \left(
\sum_{r \in \{I_{\text{T-A}}, I_{\text{F}}, I_{\text{M}} \}} w^{\text{vel}}_r
\| \text{d}x_f[r] - \text{d}\hat{x}_f[r] \|_1 \right),
\end{aligned}
\label{eq:l_recon_v3}
\end{equation}
\noindent where \(I_{\text{M}}\) denotes the subset of mouth joint indices in the skeleton, and 
\(\mathrm{d}x_f[r] = x_{f+1}[r] - x_f[r]\) represents the discrete joint velocity of region \(r\) between frames \(f\) and \(f+1\). We make sure to keep $\frac{\sum_r w_r}{\beta}$ equal across variants to preserve the relative balance between KL divergence and reconstruction loss.
The values of the scaling factors are detailed in Appendix A.1.
This objective allows to emphasize the reconstruction of certain regions such as fingers or mouth that need fine articulation for sign language, and explicitly encourages temporally more coherent motion through velocity terms.

\paragraph{Variant 4: Region-wise latent distributions (\textit{FactorVAE}).}
The final variant is based on \textit{MultiObjVAE} but instead of predicting a single latent distribution for the entire body, we predict a separate latent distribution for each region:
\begin{equation*}
q_\phi(z_{1:F}|x_{1:F}) =
\prod_{r \in \mathcal{R}}
q_{\phi_r}(z_{1:F}[r]|x_{1:F}[r]).
\end{equation*}
Latent poses $z_{1:F}$ are thus obtained by separately sampling $z_{1:F}[r] \sim \mathcal{N}(\mu_r(x_{1:F}[r]), \sigma_r(x_{1:F}[r])^2 \textbf{I})$ and concatenating them. This design aims to increase per-region reconstruction accuracy and to have more disentangled latent representations.

\paragraph{Summary.}
Table \ref{tab:vae_variants} summarizes the main differences between the four variants in terms of architecture, reconstruction objectives, and latent distribution design. Further implementation details are given in Section \ref{sec:experiments}.

\begin{table}[H]
\centering
\footnotesize
\setlength{\tabcolsep}{4pt}
\begin{tabular}{rccc}
\toprule
\textbf{} & \textbf{Enc-Dec} & \textbf{Recon. Loss} & \textbf{Distrib.}\\
\midrule
\textit{BaseVAE} & MLP & Position $\ell_1$ & Shared \\
\textit{StructVAE} & GCN + ResTempConv & Position $\ell_1$ & Shared \\
\textit{MultiObjVAE} & GCN + ResTempConv & Multi-objective & Shared \\
\textit{FactorVAE} & GCN + ResTempConv & Multi-objective & Per-region \\
\bottomrule
\end{tabular}
\caption{Summary of our four VAE variants.}
\label{tab:vae_variants}
\end{table}

\subsection{Metrics}

\subsubsection{SLP Metrics}

To evaluate both reconstructed poses from the VAE and generated sequences from the latent diffusion model, we rely on metrics commonly used in SLP, focusing on two complementary perspectives: semantic consistency and geometric accuracy of generated sign pose sequences.

\paragraph{Back-translation metrics.}

Semantic consistency is evaluated through a back-translation protocol. A pretrained sign-to-text model translates the predicted pose sequences back into text, which is then compared with the ground-truth sentence.

To ensure standardized evaluation, we use the back-translation pipeline of the SLRTP CVPR 2025 Challenge \cite{Walsh_2025_CVPR}. For clarity, we report only BLEU-1 and BLEU-4 scores~\cite{papineni2002bleu} for SLP evaluation, which measure respectively unigram and 4-gram overlaps between the predicted and reference sentences.

\paragraph{Geometric metrics.}

As in the SLRTP challenge \cite{Walsh_2025_CVPR}, motion accuracy is evaluated using the Mean Joint Error (MJE), defined as the average Euclidean distance between predicted and ground-truth joint positions:

\begin{equation}
\text{MJE} =
\frac{1}{N F}
\sum_f
\sum_j
\| x_f[j] - \hat{x}_f[j] \|_2 .
\end{equation}

In addition to joint-level errors, we introduce region-level metrics measuring the accuracy of bone orientations for the torso/arms and hands. Orientations errors enable better comparison across regions since they are independent of bone length. For a bone defined by joints $(i,j)$, the bone vector at frame $f$ is
$b_f^{(i,j)} = x_f[i] - x_f[j]$. The mean bone orientation error for region $r$ (denoted $\text{BOE}_r$) is defined as

\begin{equation}
\text{BOE}_r =
\frac{1}{|B_r| F}
\sum_f
\sum_{(i,j) \in B_r}
\left\|
\frac{b_f^{(i,j)}}{\|b_f^{(i,j)}\|_2}
-
\frac{\hat{b}_f^{(i,j)}}{\|\hat{b}_f^{(i,j)}\|_2}
\right\|_2 ,
\end{equation}

\noindent where $B_r$ is the set of parent-child joint pairs defining the bones in region $r$.

Before computing geometric metrics, the predicted sequence is truncated to the reference sequence length, and both sequences are temporally aligned using Dynamic Time Warping (DTW), so that small timing shifts do not penalize otherwise correct signing motion. DTW alignment paths are computed using either joint position distances (for MJE) or bone orientation distances (for BOE).

\subsubsection{Latent Space Analysis}

To better understand the properties of the latent space learned by our different VAE variants, we introduce the following metrics.

\paragraph{Temporal smoothness.}
Smooth latent trajectories may promote temporal coherence, but higher overall temporal variation may also reflect richer motion encoded in the latent space and preserve motion patterns relevant for sign language. Therefore, we choose to measure first- and second-order temporal variations of latent poses, denoted $S_v$ and $S_a$:
\begin{equation}
S_v = \frac{1}{d_\text{lat} (F-1)} \sum_{f=1}^{F-1} \| z_{f+1} - z_f \|_2
\end{equation}
\begin{equation}
S_a = \frac{1}{d_\text{lat} (F-2)} \sum_{f=1}^{F-2} \| z_{f+2} - 2 z_{f+1} + z_f \|_2
\end{equation}

\noindent Smaller values of $S_v$ and $S_a$ correspond to temporally smoother latent sequences.

\paragraph{Latent spatial structure.}

To analyze dependencies between latent dimensions, we compute statistics over a sufficiently large set of $K$ latent pose sequences $E_{\text{lat}} = \{ z_{1:F}^{(k)} \}_k$. The empirical covariance between latent dimensions is defined as

\begin{equation*}
\text{cov}_{ij} =
\frac{1}{KF}
\sum_{k,f}
\left(z_f^{(k)}[i] - \mu_i\right)
\left(z_f^{(k)}[j] - \mu_j\right),
\end{equation*}

\noindent where $\mu_i = \frac{1}{KF} \sum_{k,f} z_f^{(k)}[i]$ denotes the empirical mean of dimension $i$.

From this covariance matrix, we measure the average dependency between latent dimensions through the mean absolute correlation

\begin{equation}
\rho =
\frac{2}{d_{\text{lat}}(d_{\text{lat}} - 1)}
\sum_{i<j}
\left|
\frac{\text{cov}_{ij}}{\sqrt{\text{cov}_{ii}\text{cov}_{jj}}}
\right|.
\end{equation}

We also compute the effective dimensionality of the latent space~\cite{eff_dim}, which can be defined as
\begin{equation}
d_{\text{eff}} =
\frac{\left( \sum_{k=1}^{d_{\text{lat}}} \lambda_k \right)^2}
{\sum_{k=1}^{d_{\text{lat}}} \lambda_k^2}.
\end{equation}
where $\lambda_k$s denote the eigenvalues of the covariance matrix $(\text{cov}_{ij})_{ij}$. This quantity represents the equivalent number of orthogonal directions explaining the same covariance structure as the full latent space, i.e. how variance is distributed across latent dimensions.

Intuitively, a higher $d_{\text{eff}}$ suggests less collapsed ("richer") representations, allowing the generative model to exploit a greater fraction of latent dimensions information content \cite{dont_blame}. On the other hand, larger $\rho$ reflects more redundancies in the representations. Such redundancies could reduce representational efficiency but potentially facilitate optimization, as learning to denoise one dimension may implicitly provide information about other correlated dimensions. Thus, we expect these two properties to have different and possibly competing effects on diffusion training and generation performance.

\section{Experiments}
\label{sec:experiments}

\subsection{Experimental Setup}

\subsubsection{Dataset}
All experiments are conducted on the \textsc{RWTH-PHOENIX-Weather 2014T} (\textsc{Phoenix14T}) dataset~\cite{Phoenix14T}, which contains 8,257 sequences of German Sign Language performed by 9 signers. We adopt the skeleton format used in the SLRTP challenge \cite{Walsh_2025_CVPR}, consisting of $N=178$ keypoints: 8 for the torso and arms, 21 per hand, and 128 for the face.

\subsubsection{Implementation Details}

\paragraph{VAE and latent diffusion architectures.}
For the structured VAE variants (\textit{StructVAE}, \textit{MultiObjVAE} and \textit{FactorVAE}), each region-specific encoder and decoder consists of two GCN layers followed by a ResTempConv block. We fix $d_{\text{lat}}=64$ for all variants, with: $d_\text{T-A}=10$, $d_\text{RH}=24$,  $d_\text{LH}=24$, and $d_\text{face}=6$. To ensure a fair comparison, all variants have a comparable number of parameters, around $700$k ($\pm 30$k).

The latent diffusion model is mainly based on Stable Diffusion \cite{Rombach_2022_CVPR} components, with three symmetric downsampling and upsampling stages using ResNet blocks, and a central bottleneck. Diffusion timestep embeddings are injected through FiLM \cite{perez2018film}. Text conditioning is applied at all resolutions using a 1D adaptation of Stable Diffusion's SpatialTransformer\footnote{\url{https://nn.labml.ai/diffusion/stable_diffusion/model/unet_attention.html}}.

\paragraph{Training protocol.} 

In the VAE training loss, $\beta$ is increased linearly from 0 to 0.1 (to reduce the risk of posterior collapse) over the first 100 epochs.

The latent diffusion model is trained on the VAE latent means, with a fixed sequence length of 256 frames. Text conditioning is performed using the last hidden states from the pretrained transformer model \texttt{FacebookAI/xlm-roberta-base}~\cite{xlm-roberta} available in the Hugging Face library\footnote{\url{https://huggingface.co/FacebookAI/xlm-roberta-base}}.

Both the VAE and the latent diffusion model are optimized using Adam. The learning rate is set to $10^{-3}$ for VAE training and $10^{-4}$ for the diffusion model, and all models are trained for 1,000 epochs. Evaluation uses the best validation checkpoint for the VAE and the final checkpoint for the diffusion model.

All experiments are run with a fixed random seed. Reported results thus correspond to a single training run and may vary across different seeds.

\paragraph{Hardware.}  
Experiments were performed on Nvidia L40S and A40 GPUs (45G).

\subsection{Results and Discussion}

\begin{figure*}[t]
\centering
\includegraphics[
    width=\textwidth,
    trim=0.5cm 0cm 0cm 0cm,
    clip
]{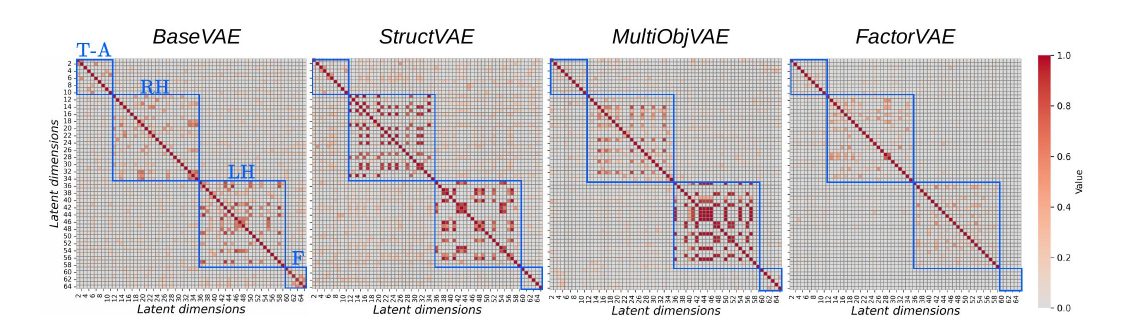}
\caption{Correlation matrices (computed from $K=256$ training sequences) between latent dimensions of VAE pose embeddings. Blue boxes delimit the different regions dimensions. T-A: torso and arms, RH: right hand, LH: left hand, and F: face.}
\label{fig:latent_corr}
\end{figure*}

\subsubsection{Impact of VAE Design on Reconstructed Poses and Latent Space}

\begin{table}[h!]
\centering
\caption{VAE reconstruction of ground truth poses on the test set: comparison between VAE variants in terms of geometric metrics.}
\label{tab:vae_variants_recon_metrics}
\begin{tabular}{rccc}
\toprule
\textbf{} & BOE\textsubscript{RH} $\downarrow$ & BOE\textsubscript{T-A} $\downarrow$ & MJE $\downarrow$ \\
\midrule
\textit{BaseVAE} & 0.28 & 0.25 & 0.012 \\
\textit{StructVAE} & 0.22 & 0.20 & 0.003 \\
\textit{MultiObjVAE} & 0.23 & 0.20 & 0.004 \\
\textit{FactorVAE} & 0.22 & 0.19 & 0.006 \\
\bottomrule
\end{tabular}
\end{table}

\begin{table}[h!]
\centering
\caption{Latent space characteristics of the VAE variants computed on the training set. $d_{\text{eff}}$ = mean value across batches. $S_v$ and $S_a$ = values averaged over all sequences.}
\label{tab:vae_variants_latent_metrics}
\begin{tabular}{rcccc}
\toprule
\textbf{} & $S_v$ & $S_a$ & $\rho$ & $d_{\text{eff}}$ \\
\midrule
\textit{BaseVAE} & 0.89 & 0.99 & 0.12 & 13 \\
\textit{StructVAE} & 0.92 & 0.88 & 0.18 & 20 \\
\textit{MultiObjVAE} & 1.07 & 1.07 & 0.13 & 22 \\
\textit{FactorVAE} & 0.87 & 0.82 & 0.05 & 18 \\
\bottomrule
\end{tabular}
\end{table}

\paragraph{Structured encoder-decoder (\textit{BaseVAE} vs \textit{StructVAE}).}

Comparing the MLP baseline (\textit{BaseVAE}) with the structured GCN+ResTempConv variant (\textit{StructVAE}), we observe consistent improvements in all geometric reconstruction metrics (Table~\ref{tab:vae_variants_recon_metrics}). In the latent space, \textit{BaseVAE} exhibits lower inter-dimension correlations ($\rho=0.12$ vs $0.18$) but a significantly lower effective dimensionality (13 vs 20). This suggests that structured encoders allow the latent space to encode more diverse and orthogonal features, possibly as skeletal connections are already modeled by the architecture, which reduces the need to encode them in the latent dimensions. 
Regarding temporal smoothness, while \textit{BaseVAE} shows slightly lower velocity $S_v$, \textit{StructVAE} achieves notably lower acceleration $S_a$ (0.88 vs 0.99), likely due to temporal convolutions accounting for longer-range motion dependencies.

\paragraph{Multi-objective loss (\textit{StructVAE} vs \textit{MultiObjVAE}).}

Introducing a multi-objective reconstruction loss in \textit{MultiObjVAE} does not significantly change geometric reconstruction metrics but impacts latent space. More precisely, $S_v$ and $S_a$ increase, mean correlation $\rho$ decreases (0.13 vs 0.18), and effective dimensionality rises (22 vs 20). This indicates that the additional loss terms and different weights encourage the model to distribute information across latent dimensions, possibly to encode finer motion patterns to minimize the error terms with higher weights on mouth, fingers and velocities.

\paragraph{Per-region latent distribution (\textit{MultiObjVAE} vs \textit{FactorVAE}).}

Shifting from a shared latent distribution (\textit{MultiObjVAE}) to per-region latent distributions (\textit{FactorVAE}) slightly improves per-region bone orientation errors (BOE\textsubscript{RH} 0.22 vs 0.23, BOE\textsubscript{T-A} 0.19 vs 0.20), but overall MJE is higher. We also observe as expected that \textit{FactorVAE}'s latent space exhibits the lowest inter-dimension correlation ($\rho=0.05$), highlighting the decoupling of regions. While this aligns with improved local reconstruction, it may come at the expense of temporal coordination across regions, as reflected in the higher MJE.

\paragraph{Latent correlation matrix insights.}

Figure~\ref{fig:latent_corr} is coherent with the above observations. We also note that across all variants, intra-region correlations dominate, especially in hands. Other notable patterns include:
\begin{itemize}
    \item in \textit{StructVAE} and \textit{MultiObjVAE}, left- and right-hand correlations show distinct structures, which reflects the predominance of right-handed signers in the dataset and the the asymmetric roles of the hands in DGS.
    \item in \textit{FactorVAE}, we observe significantly lower correlations overall, confirming reduced inter-region coordination. We also detect a posterior collapse for face embeddings, with near-zero diagonal values, which implies that the face decoder ignores these latents. One possible remedy could be to reduce the KL regularization weight for the face term in the loss \cite{fix_brok_elbo}.
\end{itemize}

\subsubsection{Trends between VAE Latent/Reconstruction Metrics and Latent Diffusion SLP}

\paragraph{Training convergence.}

\begin{figure}[hbt!]
\centering
\includegraphics[width=\linewidth]{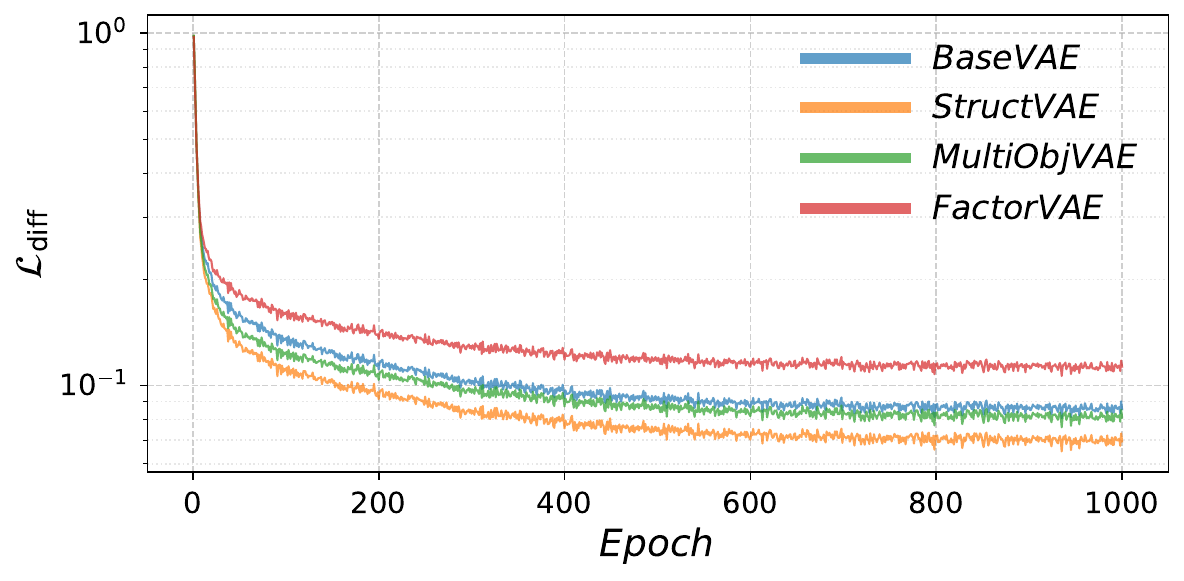}
\caption{Latent diffusion model training loss curve for each VAE variant.}
\label{fig:training_curves}
\end{figure}

Figure~\ref{fig:training_curves} compares latent diffusion training curves on different VAE latent spaces.

Interestingly, convergence speed roughly follows the inter-dimension correlation ranking $\rho$: higher correlations matches faster convergence, and variants with similar $\rho$ (\textit{BaseVAE} and \textit{MultiObjVAE}) exhibit close curves. This trend aligns with the intuition that denoising one correlated latent dimension can help improve the others, thereby accelerating training.

\paragraph{SLP back-translation vs VAE metrics.}

For reference regarding metric scales, Table~\ref{tab:baseline_bleu} reports BLEU scores obtained by training publicly available implementations of two SLP models and evaluating them using our pipeline: Progressive Transformers\footnote{https://github.com/BenSaunders27/ProgressiveTransformersSLP}~\cite{saunders_prog_trans} (2020, autoregressive) and Sign-IDD\footnote{https://github.com/NaVi-start/Sign-IDD}~\cite{tang_signidd} (2025, non-latent diffusion). We emphasize that our goal is not to achieve state-of-the-art performance, but rather to analyze how the design of the VAE affects relative variations in latent diffusion-based SLP when using a standard latent diffusion model.

\begin{table}[h]
\centering
\begin{tabular}{lcc}
\hline
Model & BLEU-1 & BLEU-4 \\
\hline
Progressive Transformers$^\dag$ \cite{saunders_prog_trans} & 33.2 & 12.1 \\
Sign-IDD \cite{tang_signidd} & 32.7 & 11.7 \\
\hline
\end{tabular}
\caption{BLEU scores of re-trained publicly available SLP models.. 
$^\dag$The original loss computation is slightly modified to average the MSE over valid frames only (excluding padding).}
\label{tab:baseline_bleu}
\end{table}

\begin{figure}[hbt!]
\centering
\includegraphics[width=\linewidth]{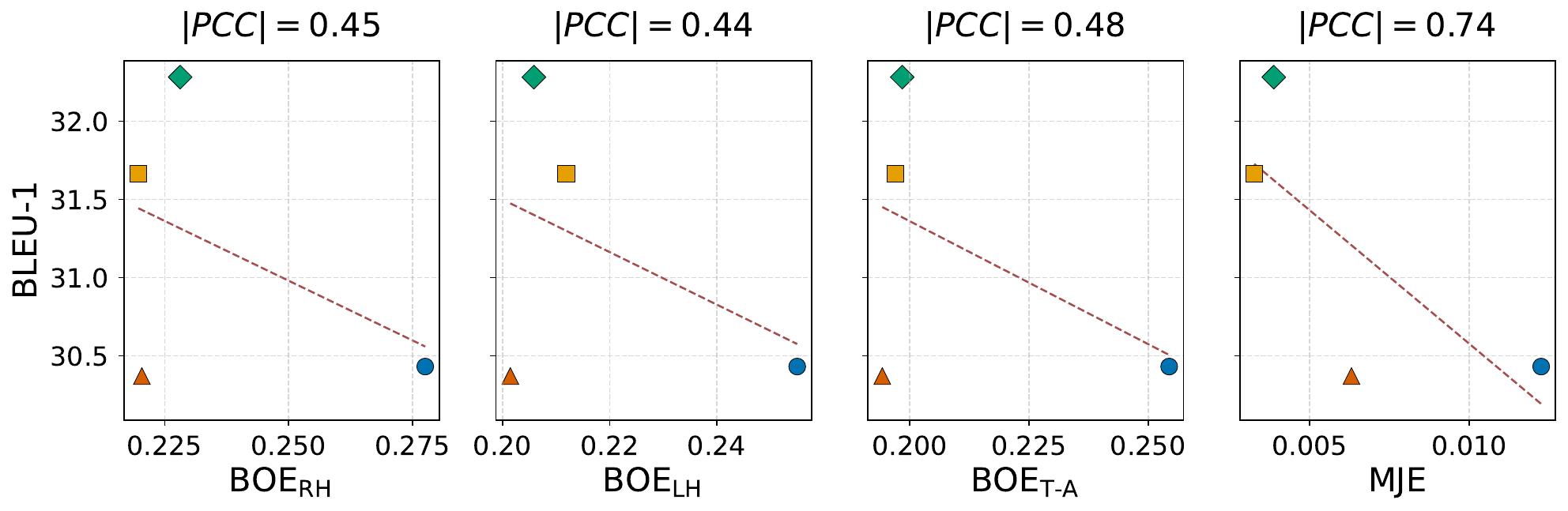}
\includegraphics[width=\linewidth]{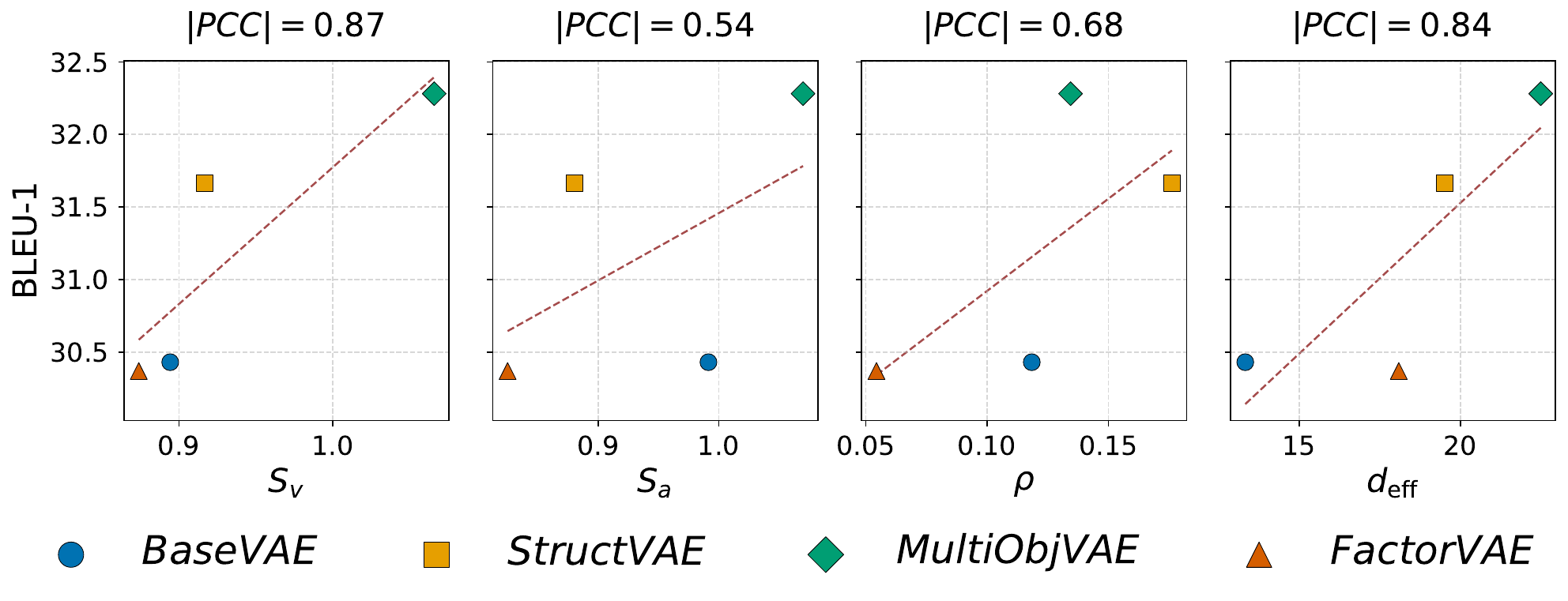}
\caption{Comparison of SLP BLEU-1 scores on text-to-sign generated poses (test set) with VAE reconstruction metrics (top) and latent space metrics (bottom). A linear fit and the absolute Pearson correlation coefficient (PCC) are reported, mainly to highlight the trend, given the limited number of observations.}
\label{fig:bleu1_vs_recon_latent}
\end{figure}

\begin{figure}[hbt!]
\centering
\includegraphics[width=\linewidth]{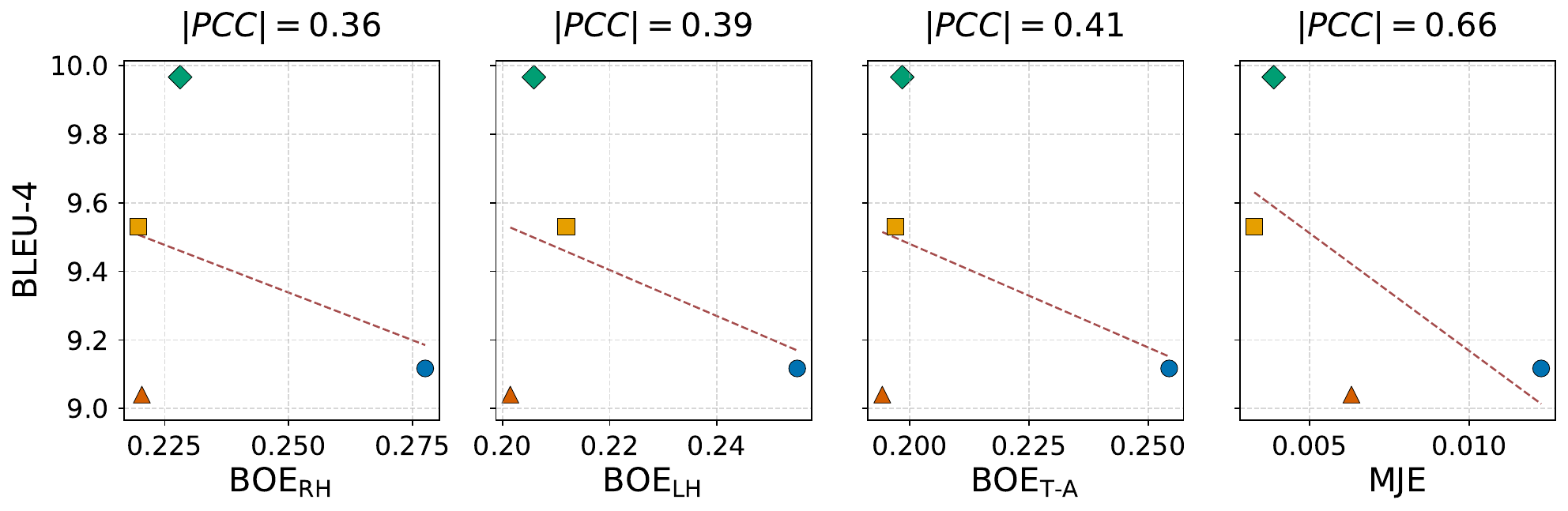}
\includegraphics[width=\linewidth]{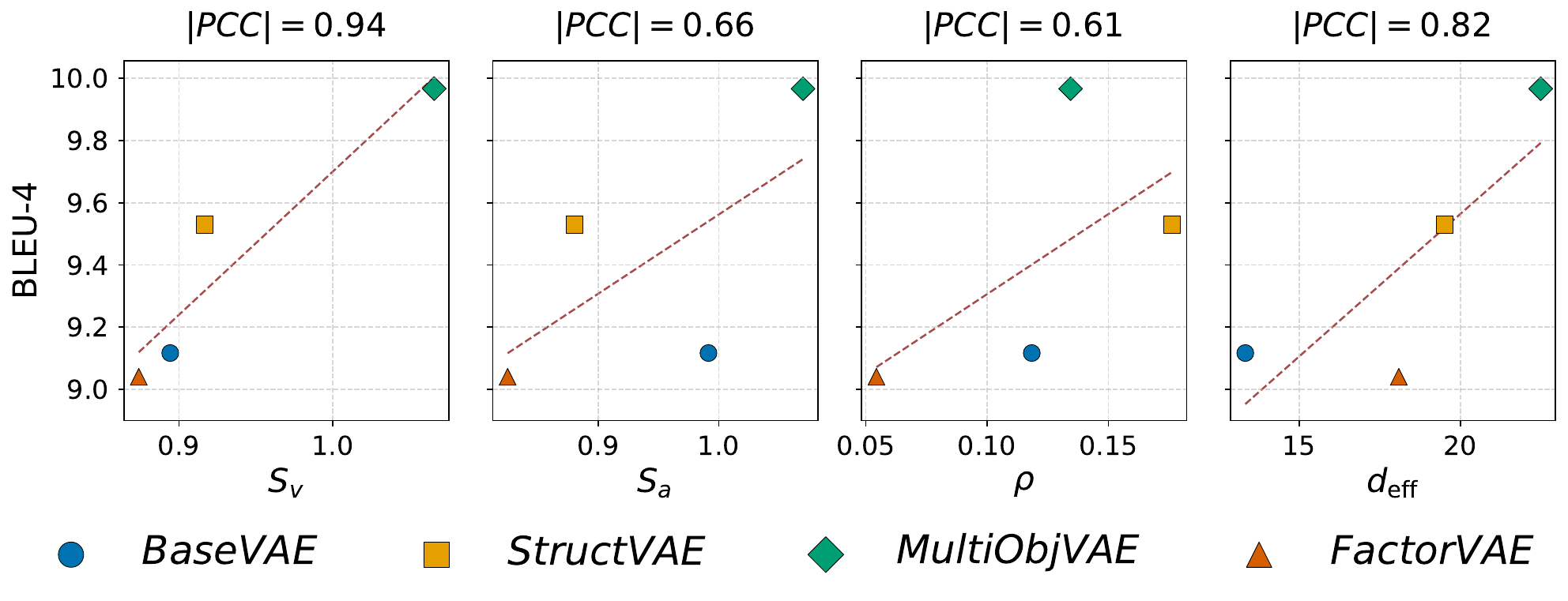}
\caption{Same comparison as Figure \ref{fig:bleu1_vs_recon_latent} using BLEU-4.}
\label{fig:bleu4_vs_recon_latent}
\end{figure}

To this extent, figures~\ref{fig:bleu1_vs_recon_latent} and~\ref{fig:bleu4_vs_recon_latent} show the BLEU-1 and BLEU-4 metrics on our latent diffusion model results (text-to-sign task) against VAE reconstruction and latent space metrics. Although the four observations are insufficient to establish statistically significant correlations, the emerging trends remain informative to assess monotonic relationships and their consistency with theoretical expectations.

In this perspective, monotonic trends are most evident when comparing BLEU to latent velocity $S_v$ or effective dimensionality ($d_{\text{eff}}$), whereas less readable when comparing with geometric reconstruction errors (as also suggested by $|\text{PCC}|<0.5$ for BOE and $|\text{PCC}|<0.75$ for MJE). For instance, large relative differences in reconstruction errors between \textit{FactorVAE} and \textit{BaseVAE} (+100\% MJE, +32\% BOE) do not translate into noticeable differences in generative performance, with BLEU-1 remaining nearly identical (30.37 vs 30.43, -0.2\%).

Besides, the observed tendency for higher latent velocity to coincide with better BLEU scores suggests that too smooth latent trajectories are less beneficial for the training of the latent diffusion SLP model, as temporal variations important for linguistic expressiveness may be absent in the latent space.\\

These observations highlight that some VAE latent space properties such as temporal variations or effective dimensionality might be more predictive of latent diffusion training behaviour and generation quality than raw VAE reconstruction accuracy. Consequently, when designing VAE for SLP, this metrics could be prioritize over marginal gains in per-joint reconstruction.

\section{Conclusion}
\label{sec:conclusion}

We presented four VAE designs for encoding sign language poses, and studied their latent space properties and reconstruction performance, as well as downstream latent diffusion-based SLP performance on a standard DGS dataset. Although VAE reconstruction errors were low and generally comparable, the learned latent spaces exhibited visible differences in temporal variation, inter-dimensional correlations, and variance allocation across dimensions. These variations were reflected in the generation quality of the diffusion model, underscoring the role of latent space design in SLP beyond the classical minimization of VAE reconstruction error.

Future work includes extending this study to larger datasets and exploring additional VAE variants to further strengthen and explore the relationship between latent sign pose space properties and SLP performance.

\section{Acknowledgements}

This work is part of \textit{Défi Inria COLaF}, financed by \textit{Plan National de Recherche en Intelligence Artificielle}. Experiments were carried out using the Grid'5000 testbed (see details at \url{https://www.grid5000.fr}).

\newpage
{
    \small
    \bibliographystyle{ieeenat_fullname}
    \bibliography{paper}
}

\clearpage
\appendix
\section{Supplementary Material}
\subsection{Weighting Factors in the Reconstruction Loss of VAE Variants}
\label{app:recon_loss_weights}

\begin{table}[H]
\centering
\begin{tabular}{lcc}
\toprule
\textbf{Name} & \multicolumn{2}{c}{\textbf{Value}} \\
\cmidrule(lr){2-3}
 & \textit{BaseVAE} / \textit{StructVAE} & \textit{MultiObjVAE} / \textit{FactorVAE} \\
\midrule
$w^{\text{pos}}_{\text{T-A}}$ & 14.5 & 5 \\
$w^{\text{pos}}_{\text{RH}}$ & 14.5 & 10 \\
$w^{\text{pos}}_{\text{LH}}$ & 14.5 & 12 \\
$w^{\text{pos}}_{\text{F}}$ & 14.5 & 4 \\
$w^{\text{pos}}_{\text{M}}$ & 0 & 8 \\
\midrule
$w^{\text{vel}}_{\text{T-A}}$ & 0 & 3 \\
$w^{\text{vel}}_{\text{F}}$ & 0 & 6 \\
$w^{\text{vel}}_{\text{M}}$ & 0 & 10 \\
\bottomrule
\end{tabular}
\caption{Weights used in $\mathcal{L}_{\text{recon}}$ depending on the VAE variant (Equations 2 and 3).}
\label{tab:vae_params}
\end{table}

\subsection{Qualitative Comparison of Reconstructed and Generated Sign Pose Sequences}

\begin{figure*}[t]
\centering

\hspace*{-2cm}
\includegraphics[width=1.15\textwidth]{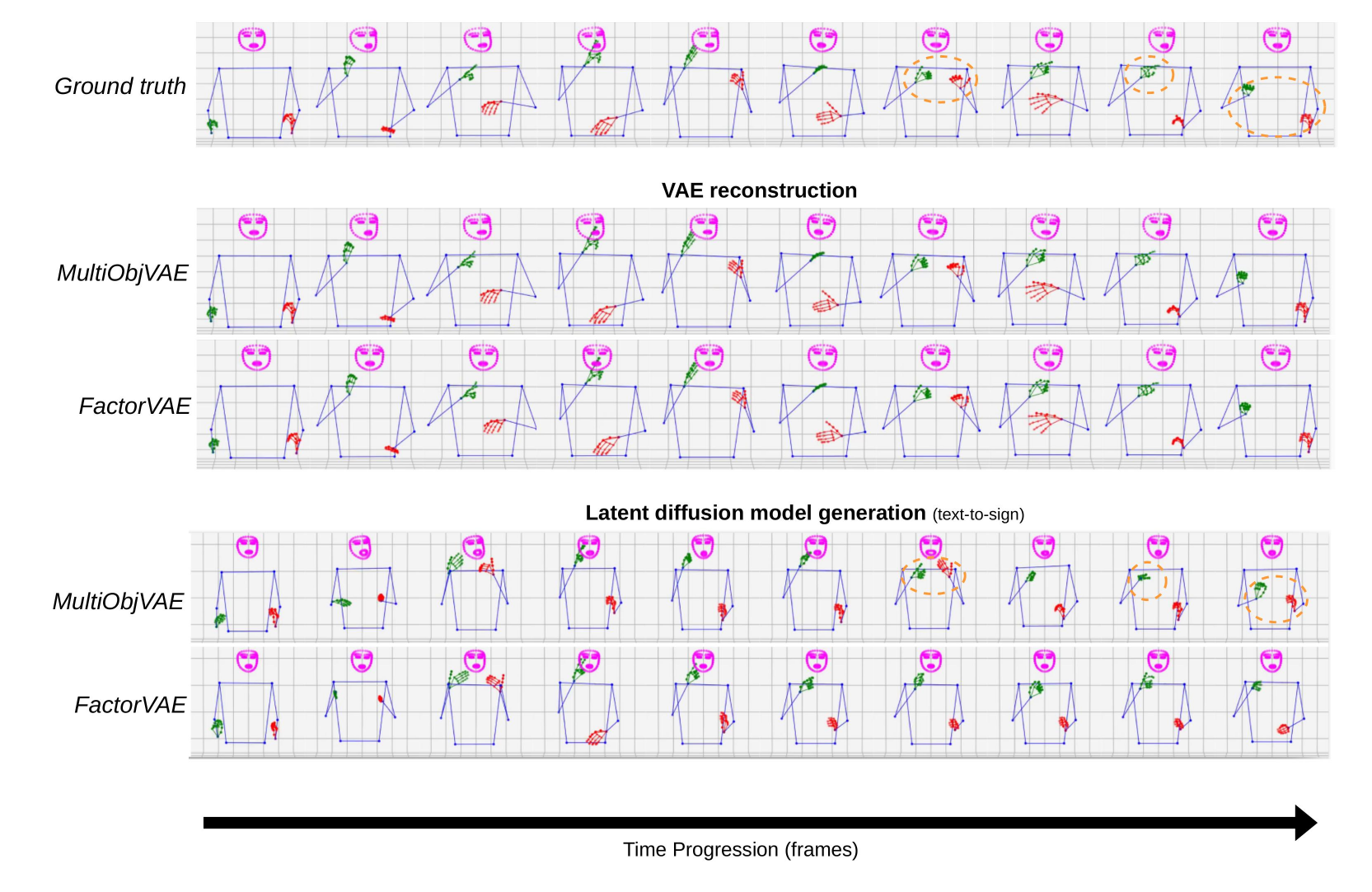}

\caption{Qualitative comparison of sign pose sequences. 
Top: ground-truth poses and VAE reconstructions produced by \textit{MultiObjVAE} and \textit{FactorVAE}. 
Bottom: text-to-sign sequences generated by latent diffusion models trained on the corresponding latent spaces. Orange doted circles indicate hand configurations correctly generated in the \textit{MultiObjVAE} latent space but not in the \textit{FactorVAE} case. \\
Sequence ID in the \textsc{Phoenix14T} dataset: \texttt{25November\_2009\_Wednesday\_tagesschau-7666}.}
\label{fig:appendix_qualitative}
\end{figure*}

Figure \ref{fig:appendix_qualitative} presents: 1) reconstructed poses from a ground-truth sign pose sequence of the \textsc{Phoenix14T} dataset using \textit{MultiObjVAE} and \textit{FactorVAE}, and 2) the generated sequences from the corresponding input sentence\footnote{\textit{"Und nun die Wettervorhersage für morgen, Donnerstag, den sechsundzwanzigsten November."} meaning \textit{"And now the weather forecast for tomorrow, Thursday, the twenty-sixth of November."}} using the latent diffusion model trained on each VAE latent space. Despite visually similar VAE reconstructions for both variants (notably with correct hands and arms motions), we observe meaningful differences between generated sequences.

\end{document}